\documentclass{article}

\usepackage[preprint]{neurips_2025}

\usepackage[utf8]{inputenc} 
\usepackage[T1]{fontenc}    
\usepackage{url}            
\usepackage{booktabs}       
\usepackage{amsfonts}       
\usepackage{nicefrac}       
\usepackage{microtype}      
\usepackage[table]{xcolor}
\usepackage{xspace}
\usepackage{graphicx}       
\usepackage{amssymb}        
\definecolor{MyDarkBlue}{rgb}{0,0.08,1}
\usepackage{subcaption}     
\definecolor{citecolor}{HTML}{0071bc}
\definecolor{shadecolor}{rgb}{0.94,0.94,0.94}
\usepackage[colorlinks, linkcolor=red, colorlinks, anchorcolor=blue, citecolor=citecolor, pagebackref=True]{hyperref}

\usepackage{array}
\newcolumntype{x}[1]{>{\centering\arraybackslash}p{#1}}
\newcolumntype{y}[1]{>{\raggedright\arraybackslash}p{#1}}
\newcolumntype{z}[1]{>{\raggedleft\arraybackslash}p{#1}}
\newcommand{\tablestyle}[2]{\setlength{\tabcolsep}{#1}\renewcommand{\arraystretch}{#2}\centering}

\usepackage{makecell}

\makeatletter
\DeclareRobustCommand\onedot{\futurelet\@let@token\@onedot}
\def\@onedot{\ifx\@let@token.\else.\null\fi\xspace}

\def\ie{\emph{i.e}\onedot}

\makeatother

\usepackage{CJKutf8}        
\usepackage{multirow}
\usepackage{amsmath}        
\usepackage{cleveref}       
\usepackage{comment}

\title{VLN-R1: Vision-Language Navigation via Reinforcement Fine-Tuning}

\author{%
  Zhangyang Qi\textsuperscript{$1,2$} \quad 
  Zhixiong Zhang\textsuperscript{$2$} \quad
  Yizhou Yu\textsuperscript{$1$} \quad
  Jiaqi Wang\textsuperscript{$2*$} \quad
  Hengshuang Zhao\textsuperscript{$1*$} \quad
  \vspace{0.1cm} \\
  \textsuperscript{$1$}The University of Hong Kong \quad
  \textsuperscript{$2$}Shanghai AI Lab \quad
  \vspace{0.1cm} \\
  {\tt\small \url{https://vlnr1.github.io}} \quad {\small * corresponding author}
  \vspace{-0.2cm}
}

\begin{document}
\begin{CJK}{UTF8}{gbsn}

\maketitle

\begin{abstract}
Vision-Language Navigation (VLN) is a core challenge in embodied AI, requiring agents to navigate real-world environments using natural language instructions. Current language model-based navigation systems operate on discrete topological graphs, limiting path planning to predefined node connections. We propose \textbf{VLN-R1}, an end-to-end framework that leverages Large Vision-Language Models (LVLM) to directly translate egocentric video streams into continuous navigation actions, adopting GRPO-based training inspired by DeepSeek-R1. To enable effective training, we first construct the \textbf{VLN-Ego} dataset using a 3D simulator, \ie, Habitat, and propose \textbf{Long-Short Memory Sampling} to balance historical and current observations. While large language models can supervise complete textual instructions, they lack fine-grained action-level control.
Our framework employs a two-stage training approach: \textbf{a)} Supervised fine-tuning (SFT) to align the model's action sequence text predictions with expert demonstrations, followed by \textbf{b)} Reinforcement fine-tuning (RFT) enhanced with a \textbf{Time-Decayed Reward (TDR)} mechanism that strategically weights multi-step future actions.
Experimental results show VLN-R1 achieves strong performance on VLN-CE benchmark. VLN-R1 proves LVLMs can drive embodied navigation and enhance task-specific reasoning through data-efficient, reward-driven post-training.
\end{abstract}

\section{Introduction}
\label{sec_Introduction}
Embodied AI is revolutionizing robotics by equipping agents with egocentric perception and physical action capabilities~\cite{evaluation_embodied, embodied-vision-language-survey, embodied-ai-survey-2, embodied-ai-survey-3}. Vision-and-language navigation (VLN)~\cite{VLN} is a fundamental challenge in Embodied AI, where agents must interpret natural language instructions to navigate in unfamiliar 3D environments. This task demands not only linguistic comprehension but also real-time decision-making in the scene.

Despite significant progress in VLN, current research still faces challenges. Existing approaches typically employ graph-based representations with fixed connectivity, constraining agents' ability to generalize to unseen or continuous environments~\cite{NaviLLM, VLMap, Discuss-before-moving, Navgpt-2, MapGPT}. 
While several studies have generalized VLN to continuous spaces (VLN-CE), replacing discrete waypoints with fine-grained motor controls to achieve free movement in simulators~\cite{CMA, Waypoint, Sim-2-Sim}，they often require additional information, like depth maps and navigation maps, and depend on specialized models like CLIP for vision-language alignment~\cite{CLIP-Nav}, limiting generalizability in human-agent interactions.

\begin{figure*}[!t]
  \vspace{-3mm}
  \centering
  \includegraphics[width=\textwidth]{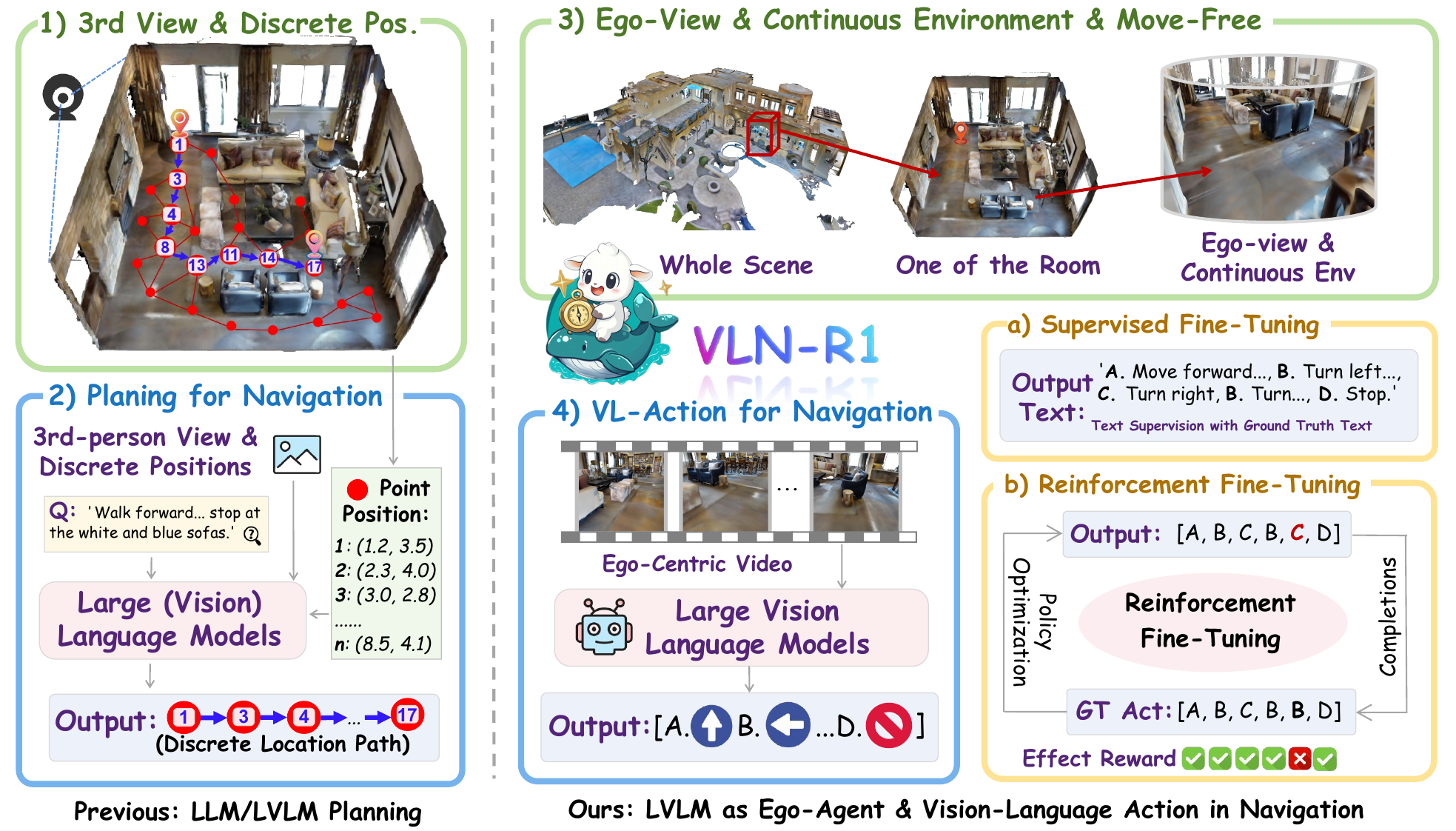}
  \vspace{-4mm}
  \caption{
      \textbf{Overview of VLN-R1.} Previous LLM/LVLM models were based on discrete positions and used a third-person perspective for path planning. In contrast, VLN-R1 directly explores in a continuous environment using first-person perspective videos. We train the LVLM using Supervised Fine-Tuning (SFT) and Reinforcement Fine-Tuning (RFT).
    }
  \label{fig_fig_teaser}
  \vspace{-4mm}
\end{figure*}

Large language models (LLM) have demonstrated impressive generalization in fields like chatbots and autonomous driving, influencing embodied intelligence.
Recently, LLMs/LVLMs have been used for vision-and-language navigation (VLN) tasks, as shown in~\Cref{fig_fig_teaser} (left). Previous methods used language models as planners, converting visual inputs into text for path planning~\cite{ANav, MiC, NavGPT, InstructNav, Discuss-before-moving, Lm-nav}. However, they still limited by these predefined navigation graphs, preventing the creation of truly embodied agents that integrate vision, language, and actions in real-world settings.
While Navid and Uni-Navi constitute important advances~\cite{Navid, Uni-NaVid}, their persistent reliance on modular visual processing pipelines fundamentally limits their scalability and generalization.

To address the limitations of LVLMs in VLN, we propose VLN-R1—a novel framework that processes egocentric video streams directly through advanced LVLMs (e.g., Qwen2-VL~\cite{Qwen2vl}) for vision-and-language navigation in continuous environments, as illustrated in part 3 of~\Cref{fig_fig_teaser}.
We first introduce \textbf{VLN-Ego}, a novel dataset for training LVLMs on continuous navigation tasks. Generated by the Habitat simulator, it comprises egocentric video streams paired with corresponding future action predictions. 
For video input processing, we implement a \textbf{Long-Short Memory Sampling} strategy that dynamically balances historical frame importance with real-time input sensitivity.

As illustrated in part 3 of ~\Cref{fig_fig_teaser}, VLN-R1 generates four fundamental action commands: \texttt{{FORWARD, TURN-LEFT, TURN-RIGHT, STOP}}. The training process consists of two stages:
\textbf{a) Supervised Fine-Tuning (SFT):} The model’s text output is fully supervised during this stage to align with the ground truth text.
\textbf{b) Reinforcement Fine-Tuning (RFT):} In this stage, we apply a reinforcement learning approach inspired by Deepseek-R1~\cite{DeepSeek-R1}, using Group Relative Policy Optimization (GRPO)~\cite{grpo} to fine-tune the model. We propose a Time-Decayed Reward (TDR) mechanism that evaluates and validates multi-step action predictions, significantly enhancing long-horizon navigation performance.

VLN-R1 achieves state-of-the-art navigation performance and enables Embodied Question Answering (EQA results are in the supplementary material)~\cite{OpenEQA}. We evaluate its navigation capabilities on continuous versions of Room-to-Room (R2R)~\cite{VLN} and Room-Across-Room (RxR)~\cite{RxR} benchmarks demonstrating the model’s effectiveness. Our paper makes these major contributions:
\begin{itemize}
\item We introduce the \textbf{VLN-Ego} dataset, a collection of egocentric video streams paired with future action predictions, specifically designed to train LVLMs for vision-and-language navigation in continuous environments.
\item We propose the \textbf{VLN-R1}, an end-to-end embodied AI framework that uses large vision-language models(LVLM) to process egocentric video streams, enabling real-time vision-and-language navigation with actions.
\item We pioneer the integration of \textbf{GRPO} and \textbf{Reinforced Fine-Tuning (RFT)} for training LVLMs on navigation tasks. Additionally, we design a Time-Decayed Reward (TDR) mechanism to specifically optimize LVLM performance in navigation scenarios.
\end{itemize}

\section{Related work}
\label{sec_Related_work}
\noindent \textbf{Vision-and-Language Navigation (VLN).}
It is a multimodal task where an agent follows natural language instructions (e.g., "Turn left, then stop by the sofa") to navigate through real or simulated visual environments. Early VLN methods~\cite{VLN, RxR, HAMT, VLN-BERT} operated in discrete environments, where agents aligned language with visual observations and teleported between predefined nodes. However, these methods struggled to generalize to open-world scenarios. The VLN-CE~\cite{vlnce} benchmark later introduced continuous environments, supporting low-level action prediction (e.g., turning angles, step lengths) and enabling free movement in simulators. Approaches include predicting low-level control commands~\cite{LAW, Sasra, AG-CMTP, CM2, WS-MGMap} or selecting subgoals generated by waypoint predictors~\cite{CMA, Waypoint, Sim-2-Sim}. Subsequent advancements leveraged vision-language pre-trained models~\cite{Uniter, Oscar, Vl-bert, Visualbert, LXMERT} or task-specific pre-training frameworks~\cite{PREVALENT, VLN-BERT, Airbert, CSAP, HOP} to enhance performance. More recently, language models (LLM) have been integrated into VLN systems~\cite{Navid, Uni-NaVid}.

\noindent \textbf{Language models as VLN agents.}
Recent advancements in Large Language Models (LLMs) have shifted vision-and-language navigation (VLN) from task-specific optimization to open-domain instruction generalization. Current methods~\cite{NaviLLM, InstructNav, VLMap, Discuss-before-moving, Navgpt-2, Lm-nav} leverage LLMs to improve multi-task navigation and cross-scene adaptability. 
They combines off-the-shelf LLMs with vision foundation models~\cite{CLIP-Nav, llava} to convert environmental observations into textual descriptions, allowing language models to guide agents. 
However, this method suffers from error accumulation and is mostly limited to discrete environments~\cite{ANav, MiC, NavGPT, InstructNav, Discuss-before-moving, Lm-nav}.
Another trend integrates large vision language models (LVLM) for continuous navigation~\cite{Navid, Uni-NaVid}, though it requires complex extra architectures. Our method uses egocentric-video-trained MLLMs within a framework similar to Deepseek-R1~\cite{DeepSeek-R1}, employing Reinforcement Fine-Tuning for training.

\noindent \textbf{Reinforcement Fine-Tuning (RFT).}
Post-training of Large Vision-Language Models (LVLMs) has traditionally used Supervised Fine-Tuning (SFT) with annotated multi-modal data for task-specific performance improvement~\cite{Qwen2vl, Llava-onevision, Deepseek-v3}. While Reinforcement Fine-Tuning (RFT) has enhanced reasoning in language models (e.g., math problem~\cite{grpo, DeepSeek-R1} and code generation~\cite{Qwen2.5-coder, preferencecode, o1-coder, Codedpo}) its application to LVLMs has been limited, mainly focusing on hallucination mitigation~\cite{llavarlhf, hadpo, povid, RlHF-V, RLAIF-V} or human preference alignment~\cite{aligning, povid}. Inspired by GRPO~\cite{grpo, GPT-4o} and Deepseek-R1-Zero~\cite{DeepSeek-R1}, which show that RL alone can improve reasoning, we pioneer the integration of verifiable reward mechanisms and GRPO-based policy optimization into the vision-and-language navigation task. It is the first application of RFT for continuous navigation decision-making in VLN.

\section{Methodology}
\label{sec_Methodology}
In this section, we introduce our methodology. \Cref{sec: Preliminary} covers RL-based post-training preliminaries and GRPO~\cite{grpo}. \Cref{sec: Dataset engine} describes our data collection pipeline from the Habitat simulator. \Cref{sec: Supervised Fine-Tuning (SFT) Stage} elaborates on the first training stage using Supervised Fine-Tuning (SFT), while \Cref{sec: Reinforcement Fine-Tuning (RFT) stage} describes the second-stage Reinforcement Fine-Tuning (RFT) process.

\subsection{Preliminary}
\label{sec: Preliminary}

\noindent \textbf{Reinforcement learning with verifiable rewards (RLVR).} It is a novel post-training approach for large language models\cite{TULU3, DeepSeek-R1, Kimi-1.5}, primarily targeting tasks with objective ground-truth answers such as mathematics~\cite{grpo} and code generation~\cite{Qwen2.5-coder, Codedpo, o1-coder}. In contrast, previous Reinforcement Learning from Human Feedback (RLHF)~\cite{Skywork-Reward, Internlm2} relies on human guidance to learn "what is correct." RLVR not only features a more concise reward function but also achieves better alignment with the intrinsic evaluation criteria of the tasks. Specifically, given an input question $q$, the policy model $\pi_{\theta}$ generates responses $o$. The optimization objective of RLVR can be formulated as:

\begin{equation}
     \max_{\pi_{\theta}} \mathbb{E}_{o \sim \pi_{\theta}(q)} \left[ R_{\text{RLVR}}(q, o) \right] = \bigg( R(q, o) - \beta \text{KL}\Big[\pi_{\theta}(o|q) \parallel \pi_{\text{ref}}(o|q)\Big] \bigg)
\end{equation}

Here, $\pi_{\text{ref}}$ denotes the reference model before optimization. Additionally, $\beta$ is the hyperparameter controlling the proportion of the KL-divergence penalty. The term $R(q, o)$ corresponds to the verifiable reward function. The specific objective function is formulated as:
\begin{equation}
    R(q, o) = 
    \begin{cases} 
    1 ,& \text{if o = ground truth}, \\
    0 ,& \text{otherwise}.
    \end{cases} 
\end{equation}

In this setup, $R(q, o)$ represents a hard reward function where the reward is set to 1 if the output $o$ matches the ground truth, and 0 otherwise. In contrast, our method introduces soft reward functions to achieve finer optimization objectives.

\noindent \textbf{R1-Zero and GRPO.}
The DeepSeek R1-Zero~\cite{DeepSeek-R1} algorithm uses Group Relative Policy Optimization (GRPO) for reinforcement learning. Unlike PPO~\cite{PPO} (which needs a critic model) or DPO~\cite{DPO} (which relies on offline preference data), GRPO~\cite{grpo} operates online without an explicit reward model, using only group-wise reward normalization, making it ideal for large language models. For a question $q$, GRPO generates $G$ responses $\{o_1, \dots, o_G\}$ from policy model $\pi_{\theta}$ and computes rewards $\{r_1, \dots, r_G\}$. It then normalizes rewards to determine relative advantages:  

\begin{equation}  
A_i = \frac{r_i - \mu \{r_1, \dots, r_G\}}{\sigma \{r_1, \dots, r_G\}},  
\end{equation}  

where $A_i$ represents the relative quality of the $i$-th answer. GRPO encourages the model to favor responses with higher reward values within the group. The optimization objective of GRPO is:  

\begin{equation}  
\scalebox{0.85}{$\displaystyle  
J_{\text{GRPO}}(\theta) = \mathbb{E} \bigg[ \sum_{i=1}^{G} \min \bigg(  
\frac{\pi_{\theta}(o_i \mid q)}{\pi_{\theta_{\text{ref}}}(o_i \mid q)} A_i,  
\quad \text{clip} \left( \frac{\pi_{\theta}(o_i \mid q)}{\pi_{\theta_{\text{ref}}}(o_i \mid q)}, 1 - \epsilon, 1 + \epsilon \right) A_i \bigg) \quad - \beta D_{\text{KL}}(\pi_{\theta} \parallel \pi_{\text{ref}}) \bigg]  
$}  
\end{equation}  

The ratio $\frac{\pi_{\theta}(o_i|q)}{\pi_{\theta_{\text{ref}}}(o_i|q)}$ quantifies preference shifts (">1" indicates higher preference for $o_i$). Advantage $A_i$ scales updates, with clipping and min operations ensuring conservative modifications: moderately boosting high-reward responses ($A_i>0$) while carefully penalizing poor ones ($A_i<0$). KL divergence prevents excessive deviation from $\pi_{\text{ref}}$.

\begin{equation}  
D_{\text{KL}} = \pi_{\text{ref}}(o_i \mid q) \left( \log \frac{\pi_{\text{ref}}(o_i \mid q)}{\pi_{\theta}(o_i \mid q)} - 1 \right)  
\end{equation}  

The KL term brings $\pi_\theta$ close to $\pi_{\text{ref}}$, avoiding reward-hacking and unnatural responses.

\begin{figure*}[!t]
  \vspace{-3mm}
  \centering
  \includegraphics[width=0.80\textwidth]{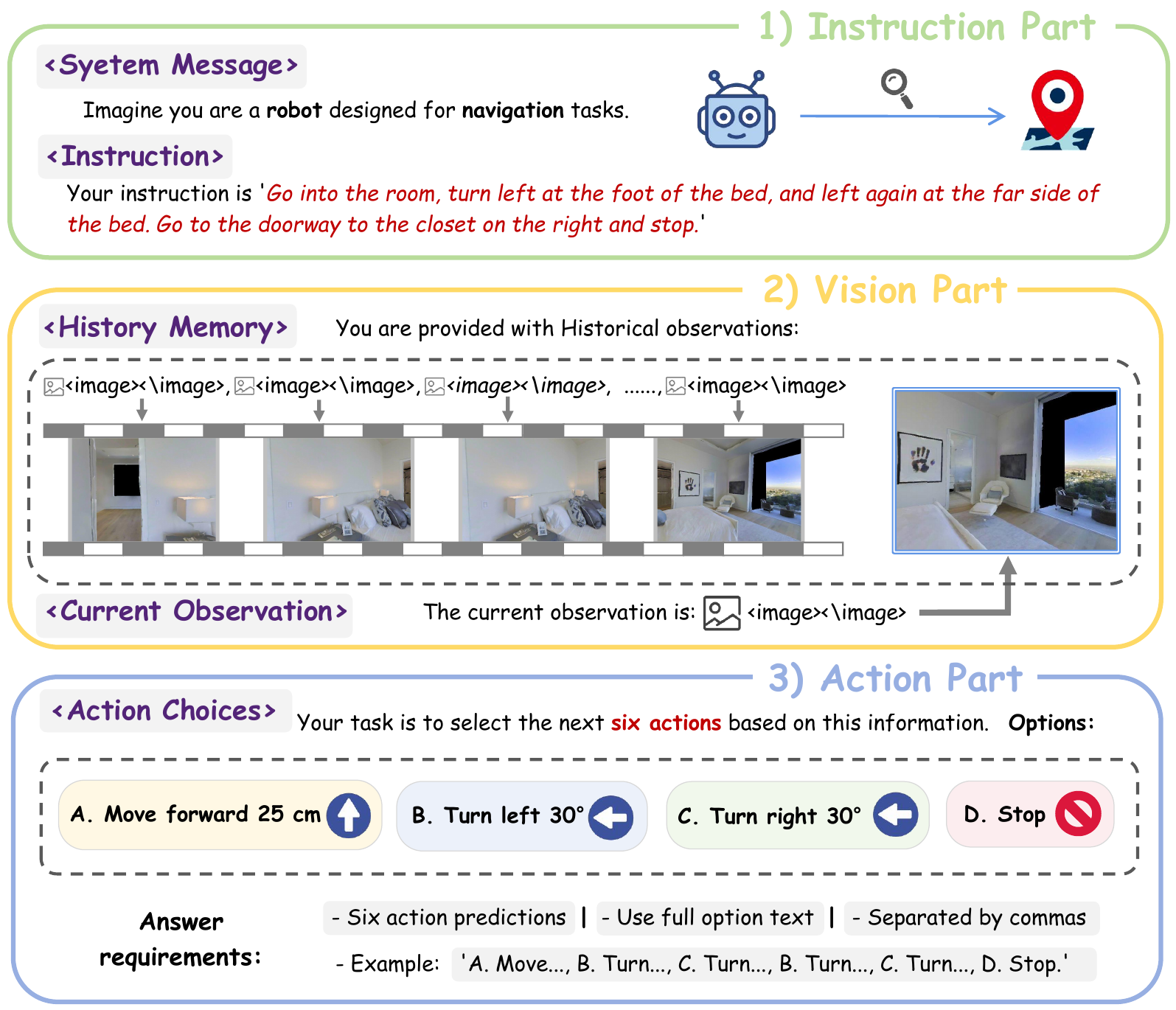}
  \caption{
      \textbf{Data Engine: VLN-Ego.} We created a dataset named VLN-Ego for LVLM-based navigation using Habitat's virtual simulation engine. Its textual annotations primarily consist of three parts: Instruction Part, Vision Part, and Action Part.
    }
  \label{fig_data_engine}
  \vspace{-4mm}
\end{figure*}

\subsection{Dataset Engine: VLN-Ego}
\label{sec: Dataset engine}
For indoor navigation tasks, previous non-LVLM approaches~\cite{wmnav, CMA, Waypoint} either operated in discrete environments or relied on additional modalities such as global maps and depth information. In contrast to prior work, our framework enables LVLM-based navigation using solely egocentric visual input, eliminating dependency on additional sensor modalities. This allows the LVLM agent to operate in a continuous environment, where it can execute flexible actions such as moving forward by specific distances or turning at specific angles. Therefore, we designed \textbf{VLN-Ego}, a first-person video-stream-based LVLM training engine for navigation.

We utilize Meta's Habitat~\cite{habitat, Habitat-3.0, vlnce} simulation platform, which is built upon the Matterport3D~\cite{Matterport3D} scene dataset. Total of 90 scenes are divided into training (61 scenes), val-seen (11 scenes), and val-unseen (18 scenes) sets, serving as the training, validation, and test sets, respectively. To maintain consistency with previous benchmarks, we also adopt the R2R~\cite{vlnce} (Room-to-Room) and RxR~\cite{RxR} (Room-across-Room) trajectories within the Matterport3D scenes. R2R primarily involves navigating from one room to another, comprising 7,189 paths, with instructions such as \textit{"Walk past the kitchen table, turn left into the hallway, and stop at the bedroom door."} RxR, on the other hand, features paths three times longer than R2R, we have selected its English-language instructions, totaling 42,023 trajectories.

VLN-Ego acquires annotations action-by-action, with the annotation structure detailed in~\Cref{fig_data_engine}. It mainly consists of three parts. The first part is the instruction part, which includes the \colorbox{green!15}{\texttt{<System Message>}} and \colorbox{green!15}{\texttt{<Instruction>}}. The latter refers to the natural language instruction provided as input. The second part is the vision part, which contains the first-person visual information, including all the historical frames \colorbox{green!15}{\texttt{<History Memory>}} and the \colorbox{green!15}{\texttt{<Current Observation>}}. The final part is the action part, which includes the \colorbox{green!15}{\texttt{<Action Choices>}}. These are the four options: \{\texttt{FORWARD}, \texttt{TURN-LEFT}, \texttt{TURN-RIGHT}, \texttt{STOP}\}. The ground truth will include the next 6 actions. Finally, we obtained 630K training samples from R2R and 1.2M training samples from RxR, all exclusively derived from the 61 training scenes.

\subsection{Supervised Fine-Tuning (SFT) Stage}
\label{sec: Supervised Fine-Tuning (SFT) Stage}

\paragraph{VLN task definition.}  
The VLN-R1 task is formally defined as follows: At each time step $t$, given a natural language instruction $\mathcal{I}$ and an egocentric video stream $\mathcal{V}_{0:t-1} = \{ v_0, v_1, \dots, v_t \}$, which consists of historical observations, the agent predicts a sequence of $n$ future actions $\{ \mathcal{A}_t, \mathcal{A}_{t+1}, \dots, \mathcal{A}_{t+n-1} \}$ in order to successfully navigate within new environments. Each action $\mathcal{A}_t$ corresponds to a low-level atomic motion. The action space consists of four primitive actions:\colorbox{gray!20}{\texttt{\{FORWARD, TURN-LEFT, TURN-RIGHT, STOP\}}}. A navigation episode is considered successful if the agent stops within a threshold distance of the target.

\paragraph{History frames selection.}
Conventional strategies for selecting historical frames have limitations. Uniform sampling weights all past frames equally, ignoring the relevance of recent observations~\cite{Waypoint}, while exponential decay over-prioritizes proximity, losing long-term context critical in Navigation~\cite{Navid}. To overcome these limitations, we propose a \textbf{Long-Short Memory Sampling} strategy, designed to balance short-term relevance with long-term contextual awareness. It is formally defined as:

\begin{equation}
\mathcal{H}_t = \underbrace{\{ v_{t-1 \cdot \delta_1}, v_{t-2 \cdot \delta_1}, \ldots, v_{t-M } \}}_{\text{Short-term memory (sampling rate $\delta_1$)}} \cup \underbrace{\{ v_{t-M-\delta_2}, v_{t-M-2\delta_2}, \ldots, v_{0} \}}_{\text{Long-term memory (sampling rate $\delta_2$)}}
\end{equation}

Short-term memory samples every $\delta_1$ frames within $M$ steps, while long-term memory samples the remaining history at $\delta_2$ interval ($\delta_2 > \delta_1$). This preserves both recent details and long-range context.

\paragraph{Prediction and loss.}  
The supervised fine-tuning (SFT) phase aligns the agent's multi-step action prediction texts with ground-truth texts. Given a historical observation sequence $\mathcal{H}_t$, instruction $\mathcal{I}$, and current observation $v_t$, the model generates an $n$-step future action sequence written as:

\begin{equation}
\hat{\mathcal{A}}_{t: \ t+n} = \mathrm{Concat}_{k=0}^{n-1} \left( \alpha_{t+k},\ \phi(\alpha_{t+k}) \right)
\end{equation}

where $\alpha_{t+k}$ denotes the action option identifier at timestep $t+k$ (e.g., $\alpha_{t}=\texttt{B}$ means selecting the second predefined action), and $\phi$ a deterministic mapping from options to action descriptions (e.g., $\phi(\texttt{B})$ yields 'Turn left 30 degrees'). The model autoregressively predicts action tokens, with the probability of the $i$-th token $w_i$ as $P(w_i | w_{1:i-1}, \mathcal{H}_t, v_t, \mathcal{I})$. We minimize the cross-entropy loss between the predicted text $\hat{\mathcal{A}}_{t:t+n-1}$ and the ground-truth $\mathcal{A}^*_{t:t+n-1}$.

\begin{equation}
\mathcal{L}_{\text{SFT}} = -\sum_{k=0}^{n-1} \sum_{j=1}^{L_k} \log P(w^*_{j} | w^*_{1:j-1}, \mathcal{H}_t, v_t, \mathcal{I})
\end{equation}

where $L_k$ denotes the token length of the $k$-th action text (e.g., "C. Turn right 30 degrees" has $L_k=6$ tokens). This loss function supervises both action option identifiers ($\texttt{A}/\texttt{B}/\texttt{C}/\texttt{D}$) and their motion descriptions. The model learns to compositionally reason by jointly generating option identifiers (discrete symbols) and linguistically coherent, context-aware action descriptions.

\begin{figure*}[!t]
  \vspace{-3mm}
  \centering
  \includegraphics[width=\textwidth]{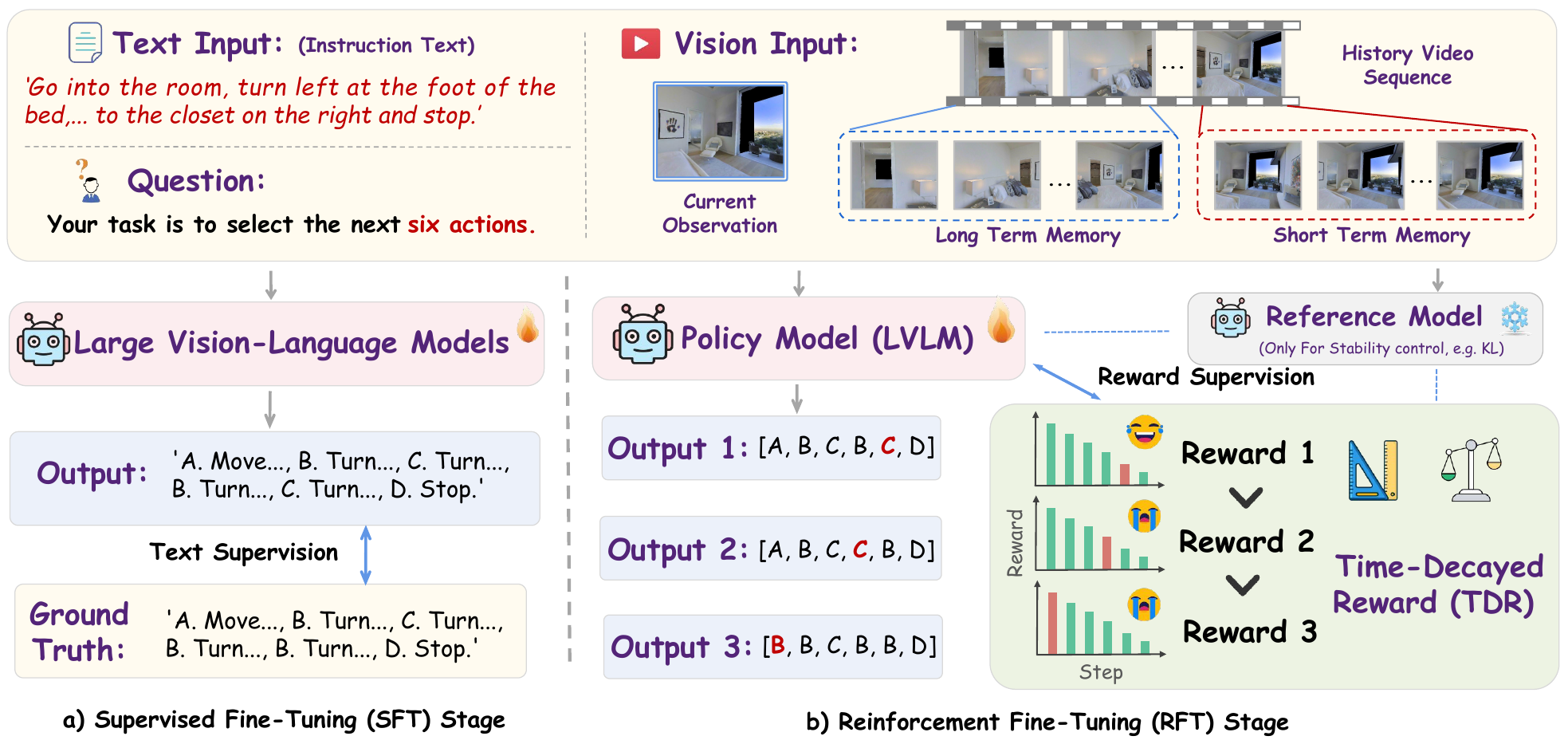}
  \caption{
      \textbf{Model Architecture of VLN-R1.} VLN-R1 employs a Long-Short Memory approach for processing visual inputs. The training consists of two stages. During the supervised fine-tuning (SFT) stage, we only supervise the output text. In the reinforcement fine-tuning (RFT) stage, we implement supervision using a designed Time-Decayed Reward (TDR) mechanism.
    }
  \label{fig_model_arch}
  \vspace{-3mm}
\end{figure*}

\subsection{Reinforcement Fine-Tuning (RFT) stage}
\label{sec: Reinforcement Fine-Tuning (RFT) stage}

After Supervised Fine-Tuning (SFT), we further optimize the model with \textbf{Reinforcement Fine-Tuning (RFT)} in the second phase. While using the same instruction data format, we design an innovative reward function. Previous reinforcement fine-tuning uses simplistic reward mechanisms~\cite{grpo, Codedpo, grpo}, such as binary option matching, error thresholds, or text similarity metrics like ROUGE-L. However, these methods are insufficient for vision-and-language navigation (VLN) tasks, as they lack effective supervision for temporal predictions.

We propose a \textbf{Time-Decayed Reward (TDR)} function to address temporal dependencies. Our method extracts the action component $\alpha_{t+k}$ from predicted text tuples $(\alpha_{t+k}, \ \phi(\alpha_{t+k}))$ to form sequential predictions: $A_{t:t+n} = (\alpha_t, \alpha_{t+1}, \dots, \alpha_{t+n-1})$ for $n$ future steps. Our temporal weighting scheme applies exponential decay $\gamma$ to action significance, ensuring proximal actions receive substantially higher consideration than distant ones. The reward is shown as below:

\begin{equation}
R_{\text{nav}} = \sum_{k=0}^{n-1} \gamma^k \cdot \mathbb{I} \ (\alpha_{t+k} = \alpha^*_{t+k})
\end{equation}

where $\mathbb{I}$ be the indicator function that outputs 1 when the predicted action matches the ground truth $\alpha^*_{t+k}$, and 0 otherwise. This design ensures position-aware consistency by applying the exponentially decaying $\gamma^k$ term, which prioritizes earlier correct actions. Our approach overcomes the limitation of supervised fine-tuning (SFT), which can only optimize the overall text output but fails to provide precise guidance for action sequence generation.

\section{Experiments}
\label{sec_Experiments}
We presents our experimental setup and results here. In \Cref{sec: Implementation details} , we explain the implementation details. \Cref{sec: Main results} shows our main results, while \Cref{sec: Ablation study} conducts the ablation study.

\subsection{Implementation Details}
\label{sec: Implementation details}

We utilize VLN-Ego as our training dataset, with the SFT phase employing 1.8M samples from both R2R~\cite{VLN} and RxR~\cite{RxR} datasets. In the RFT phase, we randomly select 10K samples from each, totaling 20K samples for training. The agent navigates to the goal using only ego-view video, and a navigation episode is successful if the agent stops within a threshold distance. We evaluate on 18 held-out scenes (Val-Unseen) using VLN-CE~\cite{vlnce} metrics. \textbf{SR$\uparrow$ (Success Rate)}, \textbf{OS$\uparrow$ (Oracle SR)}, and \textbf{SPL$\uparrow$ (Success weighted by Path Length)} measure navigation accuracy, while \textbf{NE$\downarrow$ (Navigation Error)} and \textbf{TL$\downarrow$ (Trajectory Length)} assess efficiency. All distance measurements are reported in meters (m). We train both Qwen2-VL-2B and Qwen2-VL-7B models~\cite{Qwen2vl}, with the 7B model deployed on 8 NVIDIA A800 GPUs using DeepSpeed ZeRO-3 optimization~\cite{zero}. During training, images are resized to a maximum resolution of 65,536 pixels, processed in batches of 16 images per instance (generating ~4.1K tokens). For Supervised Fine-Tuning (SFT), we employ a learning rate of 5e-6 with cosine scheduling (10\% warmup), a per-GPU batch size of 2, and achieve a global batch size of 64 through 4 gradient accumulation steps, completing 1 epoch in ~36 hours. For Reinforcement Fine-Tuning (RFT), we reduce the learning rate to 1e-6 with weight decay (0.01) and β=0.04, configure GRPO~\cite{grpo} with 8 samples per prompt, use a per-GPU batch size of 1 (no gradient accumulation), and complete training in about 12 hours per epoch.

\subsection{Main Results}
\label{sec: Main results}
Our main experiments include results from training on both R2R and RxR in VLN-CE. We evaluate on the Val-Unseen split, which contains completely unseen environments during training. Beyond demonstrating our method's effectiveness, we further investigate the role of RFT in this framework. Our results on VLN-CE R2R are shown in~\Cref{tab_vlnce_R2R}. First, our method relies solely on RGB video inputs while achieving state-of-the-art performance. The 2B model outperforms the 7B SFT results via RFT, consistent with Deepseek-R1’s findings~\cite{DeepSeek-R1} that Reinforcement Fine-Tuning (RFT) enables smaller models to match the performance of larger-size counterparts.

The results on VLN-CE RxR are shown in~\Cref{tab_vlnce_RxR}. The last four rows demonstrate training with the full dataset, where the results align with the conclusions from R2R. The middle four rows represent a configuration where supervised fine-tuning (SFT) used only 630K R2R samples, while reinforcement fine-tuning (RFT) utilized 10K RxR samples. Surprisingly, the LVLM fine-tuned with merely 10K samples via RFT outperforms its counterpart trained on the complete dataset. Remarkably, RFT enables effective cross-domain adaptation: after base training in one domain, just a small fraction of new domain data suffices for superior performance transfer. We present qualitative experimental results in~\Cref{fig_qualitative_results}, with more qualitative examples and Embodied QA results in the appendix.

\begin{table}[t]
        \centering
        \tablestyle{5pt}{1.3}
        \resizebox{0.95\linewidth}{!}{
        \begin{tabular}{l|cccc|ccccc}
\toprule
\multirow{2}{*}{Method} & \multicolumn{4}{c|}{Observation} & \multicolumn{5}{c}{VLN-CE R2R Val-Unseen} \\
                        & Map  & Odom. & Depth & RGB  & \textbf{SR}$\uparrow$ & \textbf{OS}$\uparrow$ & \textbf{SPL}$\uparrow$ & \textbf{NE}$\downarrow$ & TL \\ 
\midrule
\multicolumn{5}{l}{~~\textit{Task-Specific method}} \\
AG-CMTP~\cite{AG-CMTP}          & \checkmark & \checkmark & \checkmark &            & 23.1 & 39.2 & 19.1 & 7.90 & – \\
R2R-CMTP~\cite{AG-CMTP}          & \checkmark & \checkmark & \checkmark &            & 26.4 & 38.0 & \textbf{22.7} & 7.90 & – \\
\midrule
\multicolumn{5}{l}{~~\textit{Ego-view LVLM based method}} \vspace{1mm} \\
VLN (SFT, Qwen2-VL-2B)          &            &            &            & \checkmark & 21.2 & 33.0 & 15.9 & 8.27 & 11.9 \\
VLN (SFT, Qwen2-VL-7B)          &            &            &            & \checkmark & 24.9 & 37.1 & 17.5 & 7.92 & 15.0 \\
VLN-R1 (Qwen2-VL-2B)            &            &            &            & \checkmark & 25.6 & 37.5 & 20.5 & 10.2  & 16.8 \\
\rowcolor{green!10}
VLN-R1 (Qwen2-VL-7B)            &            &            &            & \checkmark & \textbf{30.2} & \textbf{41.2} & 21.8 & \textbf{7.0} & 10.0 \\
\bottomrule
\end{tabular}
        }
        \vspace{2mm}
        \caption{\textbf{Comparison on VLN-CE R2R Val-Unseen~\cite{VLN, vlnce}.} 'VLN (SFT)' refers to results after the first-stage supervised fine-tuning. The 2B model performs similarly to the 7B model after reinforcement fine-tuning (RFT).}
        \label{tab_vlnce_R2R}
        \vspace{-4mm}
\end{table}

\begin{table}[!t]
    \vspace{-2mm}
    \centering
    \tablestyle{5.5pt}{1.3}
    \resizebox{0.95\linewidth}{!}{
    \begin{tabular}{l|ccc|ccccc}
\toprule
\multirow{2}{*}{Method}         & \multicolumn{3}{c|}{Observation}     & \multicolumn{5}{c}{VLN-CE RxR Val-Unseen}                             \\ 
 & Odom. & Depth & S.RGB & \textbf{SR}$\uparrow$ & \textbf{OS}$\uparrow$ & \textbf{SPL}$\uparrow$ & \textbf{NE}$\downarrow$ & TL \\ 
 
\midrule
\multicolumn{5}{l}{~~\textit{Task-Specific method}} \\
LAW*~\cite{LAW}  & \checkmark & \checkmark & \checkmark & 8.0 & 21.0 & 8.0 & 10.9 & 4.0 \\
CM2*~\cite{CM2}   & \checkmark & \checkmark & \checkmark & 14.4  & 25.3 & 9.2 & \textbf{9.0} & 12.3  \\
WS-MGMap*~\cite{WS-MGMap}  & \checkmark & \checkmark & \checkmark & 15.0  & 29.8 & 12.1  & 9.8 & 10.8 \\
Seq2Seq*~\cite{vlnce} &  & \checkmark & \checkmark & 3.51  & 5.02 & 3.4 & 11.8 & 1.2  \\
CMA*~\cite{vlnce}  &  & \checkmark & \checkmark & 4.41 & 10.7         & 2.5 & 11.7 & 5.1  \\
$A^2$Nav$^\dagger$~\cite{ANav} &  & & \checkmark & 16.8  & --           & 6.3 & --  & -- \\

\midrule
\multicolumn{5}{l}{~~\textit{Ego-view LVLM based method}} \vspace{1mm} \\
VLN (SFT, Qwen2-VL-2B)*  &  &  & \checkmark & 14.1 & 22.3  & 11.2& 9.8  & 13.5  \\
VLN (SFT, Qwen2-VL-7B)*  &  &  & \checkmark & 14.9  & 23.0 & 11.9 & 10.8  & 11.9 \\
VLN-R1 (Qwen2-VL-2B)$\ddag$  &  &  & \checkmark & 20.7  & 30.1  & 16.9 & 10.2  & 12.6 \\
\rowcolor{green!10}
VLN-R1 (Qwen2-VL-7B)$\ddag$  &  &  & \checkmark & \textbf{22.7}  & 30.4  & 17.6 & 9.1  & 12.4  \\
\midrule
VLN (Qwen2-VL-2B)  &  &  & \checkmark & 18.7  & 27.4  & 16.2 & 11.2  & 18.4  \\
VLN (Qwen2-VL-7B)  &  &  & \checkmark & 19.5  & 27.5  & 16.7 & 10.6  & 15.3  \\
VLN-R1 (Qwen2-VL-2B)  &  &  & \checkmark & 21.4 & 30.6 & 15.5 & 10.2 & 15.6  \\
\rowcolor{green!10}
VLN-R1 (Qwen2-VL-7B)  &  &  & \checkmark & \textbf{22.3}  & \textbf{33.4} & \textbf{17.5} & 10.4  & 15.3 \\
\bottomrule
\end{tabular}

    }
    \vspace{2mm}
    \caption{\textbf{Comparison on VLN-CE RxR Val-Unseen~\cite{RxR, vlnce}.} $^*$: only trained on VLN-CE R2R. $\ddag$indicates that SFT was trained solely on R2R, while RFT incorporates data from RxR. Our results show that RFT quickly boosts smaller models' performance and excels in cross-domain generalization, achieving strong RxR results with minimal data after R2R pretraining.}
    \label{tab_vlnce_RxR}
    \vspace{-6mm}
\end{table} 

\subsection{Ablation study}
\label{sec: Ablation study}
We conduct a series of ablation experiments to analyze model performance during the Supervised Fine-Tuning (SFT) and Reinforcement Fine-Tuning (RFT) phases. All experiments used Qwen2-VL-7B~\cite{Qwen2vl}, trained solely on the R2R.

\paragraph{Supervised fine-tuning (SFT) stage.}
\label{sec: Ablation study on the Supervised Fine-Tuning (SFT) stage}
We investigate how different designs of the Action Space and History Memory components affect VLN-R1's performance. For these experiments, we used supervised fine-tuning (SFT) exclusively during training. As shown in \Cref{tab_aba_action_space}, our analysis reveals that continuous prediction of the next 6 actions yields optimal results. We observe significant performance degradation when predicting only single actions, as this approach fails to account for future step dependencies. \Cref{tab_aba_history_memory} demonstrates that our proposed Long-Short Memory strategy achieves the best frame selection performance by effectively balancing current observations with historical context.

\begin{figure*}[!t]
  \vspace{-7mm}
  \centering
  \includegraphics[width=1.0\textwidth]{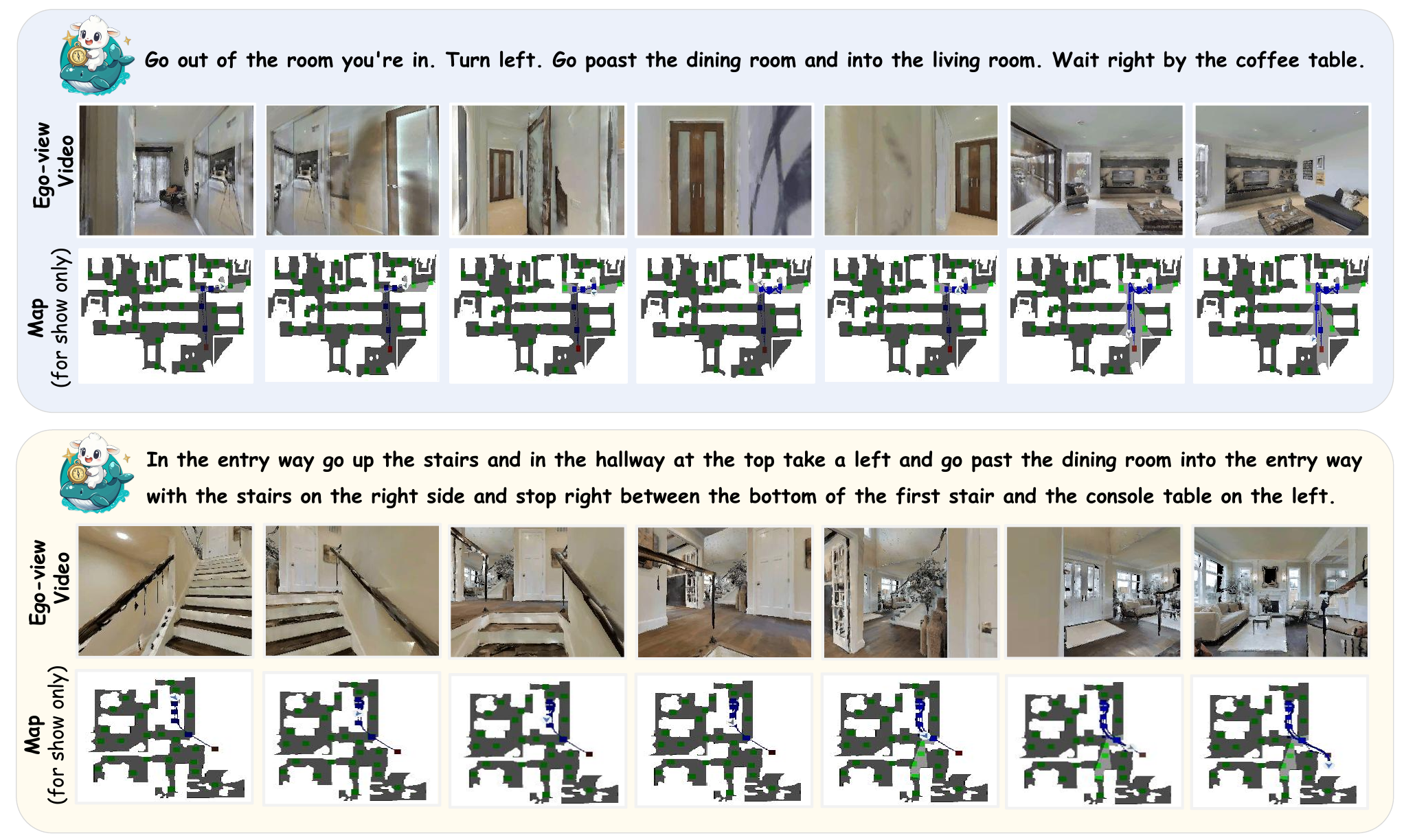}
  \vspace{-2mm}
  \caption{
      \textbf{Qualitative Results of VLN-R1.} As shown, VLN-R1 accepts egocentric video input and navigates through a continuous environment to ultimately reach the target location.
    }
  \label{fig_qualitative_results}
  \vspace{-4mm}
\end{figure*}

\begin{table*}[t]
    \label{tab:ablation_studies}
    \begin{minipage}{0.90\textwidth}
        \centering
        \begin{minipage}{0.43\textwidth}
            \centering
            \subcaption{\textbf{Action Space}}
            \tablestyle{6pt}{1.2}
            \begin{tabular}{l|cc}
\toprule
\multirow{2}{*}{\textbf{Action Space Variant}} & \multicolumn{2}{c}{R2R Val} \\ 
 & \textbf{SR}$\uparrow$ & \textbf{OS}$\uparrow$ \\ 
\midrule
Single Discrete Action       & 15.1 & 33.6 \\
4-Discrete-Action Set      & 21.4 & 29.6 \\
\rowcolor{green!10}
6-Discrete-Action Set        & 24.9 & 31.7 \\
8-Discrete-Action Set        & 22.7 & 30.4 \\
\bottomrule
\end{tabular}
            \label{tab_aba_action_space}
        \end{minipage}
        \hspace{10mm}
        \begin{minipage}{0.43\textwidth}
            \centering
            \subcaption{\textbf{History Memory}}
            \tablestyle{6pt}{1.2}
            \begin{tabular}{l|cc}
\toprule
\multirow{2}{*}{\textbf{History Memory Method}} & \multicolumn{2}{c}{R2R Val} \\ 
 & \textbf{SR}$\uparrow$ & \textbf{OS}$\uparrow$ \\ 
\midrule
Average Sampling (8)     & 20.8 & 28.9 \\
Average Sampling (16)    & 22.0 & 31.3 \\
Exponential Decay        & 23.8 & 34.3 \\
\rowcolor{green!10}
Long Short Memory        & 24.9 & 31.7 \\
\bottomrule
\end{tabular}
            \label{tab_aba_history_memory}
        \end{minipage}
        
        \begin{minipage}{0.43\textwidth}
            \centering
            \subcaption{\textbf{Number of Generations}}
            \tablestyle{6pt}{1.2}
            \begin{tabular}{l|cc}
\toprule
\multirow{2}{*}{\textbf{RFT Generations}} & \multicolumn{2}{c}{R2R Val} \\ 
 & \textbf{SR}$\uparrow$ & \textbf{OS}$\uparrow$ \\ 
\midrule
\textit{k} = 2 (warm start)   & 24.7  & 32.5 \\
\textit{k} = 4                & 26.5 & 35.4  \\
\textit{k} = 6                & 28.4  & 37.2  \\
\rowcolor{green!10}
\textit{k} = 8 (convergence)  & 30.2  & 41.2  \\
\bottomrule
\end{tabular}
            \label{tab_aba_num_of_generations}
        \end{minipage}
        \hspace{7mm}
        \begin{minipage}{0.43\textwidth}
            \centering
            \subcaption{\textbf{Reward Function}}
            \tablestyle{6pt}{1.2}
            \begin{tabular}{l|cc}
\toprule
\multirow{2}{*}{\textbf{Reward Type}} & \multicolumn{2}{c}{R2R Val} \\ 
 & \textbf{SR}$\uparrow$ & \textbf{OS}$\uparrow$ \\ 
\midrule
Hard Reward                         & 23.8  & 32.3 \\ 
Uniform (all actions equal)         & 25.0  & 33.0 \\
Linear Distance-weighting           & 28.3  & 33.6 \\ 
\rowcolor{green!10}
Exponential Decay                   & 30.2  & 41.2 \\
\bottomrule
\end{tabular}
            \label{tab_aba_reward_function}
        \end{minipage}
    \end{minipage}
    \vspace{-1mm}
    \caption{\textbf{Ablation Studies on Training Stages}}
    \vspace{-5mm}
\end{table*}

\paragraph{Reinforcement Fine-Tuning (RFT) stage.}
\label{Ablation study on the Reinforcement Fine-Tuning (RFT) stage}
We further investigate the hyperparameters in Reinforcement Fine-Tuning (RFT). First, we examine the impact of the number of generations in ~\Cref{tab_aba_num_of_generations}, followed by an exploration of different reward functions and their effects on the results. We find that the performance improvement from 6 to 8 generations is marginal, so we choose 8. ~\Cref{tab_aba_reward_function} shows the reward functions, demonstrating the effectiveness of our Time-Decayed Reward function.

\vspace{-3mm}
\section{Conclusion and Limitation}
\vspace{-3mm}
\label{sec_conclusion}
We present VLN-R1, an end-to-end framework that leverages large vision-language models (LVLMs) for continuous vision-and-language navigation (VLN) tasks. Our framework processes egocentric visual input through LVLMs, removing the need for navigation graphs or additional sensors. VLN-R1 integrates Reinforcement Fine-Tuning (RFT) with Group Relative Policy Optimization (GRPO) and Time-Decayed Reward (TDR), improving long-horizon decision-making. Our approach achieves SOTA on VLN-CE, with the 2B model matching 7B performance after RFT, showing efficient cross-domain adaptation with minimal data. Our method has limitations, including evaluation in only simulated indoor environments, limiting real-world generalization. The discrete action space also restricts fine-grained control. Nonetheless, our work advances the navigation field by integrating Qwen2-VL, establishing VLN as a downstream task for LVLMs.

\clearpage




\end{CJK} 
\end{document}